\def\year{2022}\relax
\documentclass[letterpaper]{article} 
\usepackage{aaai22}  
\usepackage{times}  
\usepackage{helvet}  
\usepackage{courier}  
\usepackage[hyphens]{url}  
\usepackage{graphicx} 
\usepackage{natbib}  
\usepackage{caption} 
\DeclareCaptionStyle{ruled}{labelfont=normalfont,labelsep=colon,strut=off} 
\usepackage{algorithm}
\usepackage{algorithmic}
\usepackage{hyperref}
\usepackage{newfloat}
\usepackage{listings}
\floatstyle{ruled}
\newfloat{listing}{tb}{lst}{}
\floatname{listing}{Listing}
\nocopyright
\usepackage{graphicx}
\usepackage{epsfig}
\usepackage{amsmath}
\usepackage{amssymb}
\usepackage{times,graphicx,tabulary,multirow,xspace,xcolor}
\usepackage[inline, shortlabels]{enumitem}
\usepackage{booktabs}

\usepackage{pifont}
\newcommand{\cmark}{\color{gray}\ding{51}}%
\newcommand{\xmark}{\color{gray}\ding{55}}%
\definecolor{darkgreen}{rgb}{0.29, 0.33, 0.13}
\definecolor{darkred}{rgb}{0.75, 0.0, 0.0}
\definecolor{darkblue}{rgb}{0.0, 0.0, 0.75}
\definecolor{LightGray}{rgb}{0.92,0.92,0.92}

\newcommand{\oursmodel}{\textbf{\texttt{PICa}}}
\newcommand{\xv}{{\boldsymbol x}}
\newcommand{\yv}{{\boldsymbol y}}
\newcommand{\hv}{{\boldsymbol h}}
\newcommand{\Ccal}{{\mathcal{C}}}

\newcommand{\eat}[1]{}

\title{An Empirical Study of GPT-3 for Few-Shot Knowledge-Based VQA}

\author {Zhengyuan Yang, Zhe Gan, Jianfeng Wang, Xiaowei Hu,\\ Yumao Lu, Zicheng Liu, Lijuan Wang}
\affiliations{Microsoft Corporation \\ 
{\{zhengyang, zhe.gan, jianfw, xiaowei.hu, yumaolu, zliu, lijuanw\}@microsoft.com}
}

\begin{document}

\maketitle

\begin{abstract}
  Knowledge-based visual question answering (VQA) involves answering questions that require external knowledge not present in the image. Existing methods first retrieve knowledge from external resources, then reason over the selected knowledge, the input image, and question for answer prediction. However, this two-step approach could lead to mismatches that potentially limit the VQA performance. For example, the retrieved knowledge might be noisy and irrelevant to the question, and the re-embedded knowledge features during reasoning might deviate from their original meanings in the knowledge base (KB).
To address this challenge, we propose~\oursmodel, a simple yet effective method that \textbf{\texttt{P}}rompts GPT-3 via the use of \textbf{\texttt{I}}mage \textbf{\texttt{Ca}}ptions, for knowledge-based VQA. Inspired by GPT-3's power in knowledge retrieval and question answering, instead of using \emph{structured} KBs as in previous work, we treat GPT-3 as an \emph{implicit} and \emph{unstructured} KB that can jointly acquire and process relevant knowledge. Specifically, we first convert the image into captions (or tags) that GPT-3 can understand, then adapt GPT-3 to solve the VQA task in a \emph{few-shot} manner by just providing a few in-context VQA examples. We further boost performance by carefully investigating: ($i$) what text formats best describe the image content, and ($ii$) how in-context examples can be better selected and used.~\oursmodel~unlocks the first use of GPT-3 for multimodal tasks. By using only $16$ examples,~\oursmodel~surpasses the supervised state of the art by an absolute $+8.6$ points on the OK-VQA dataset. We also benchmark~\oursmodel~on VQAv2, where \oursmodel~also shows a decent few-shot performance. \footnote{Code is available at \url{https://github.com/microsoft/PICa}.}
\end{abstract}

\section{Introduction}
\begin{figure}[t]
\begin{center}
  \centerline{\includegraphics[width=8.5cm]{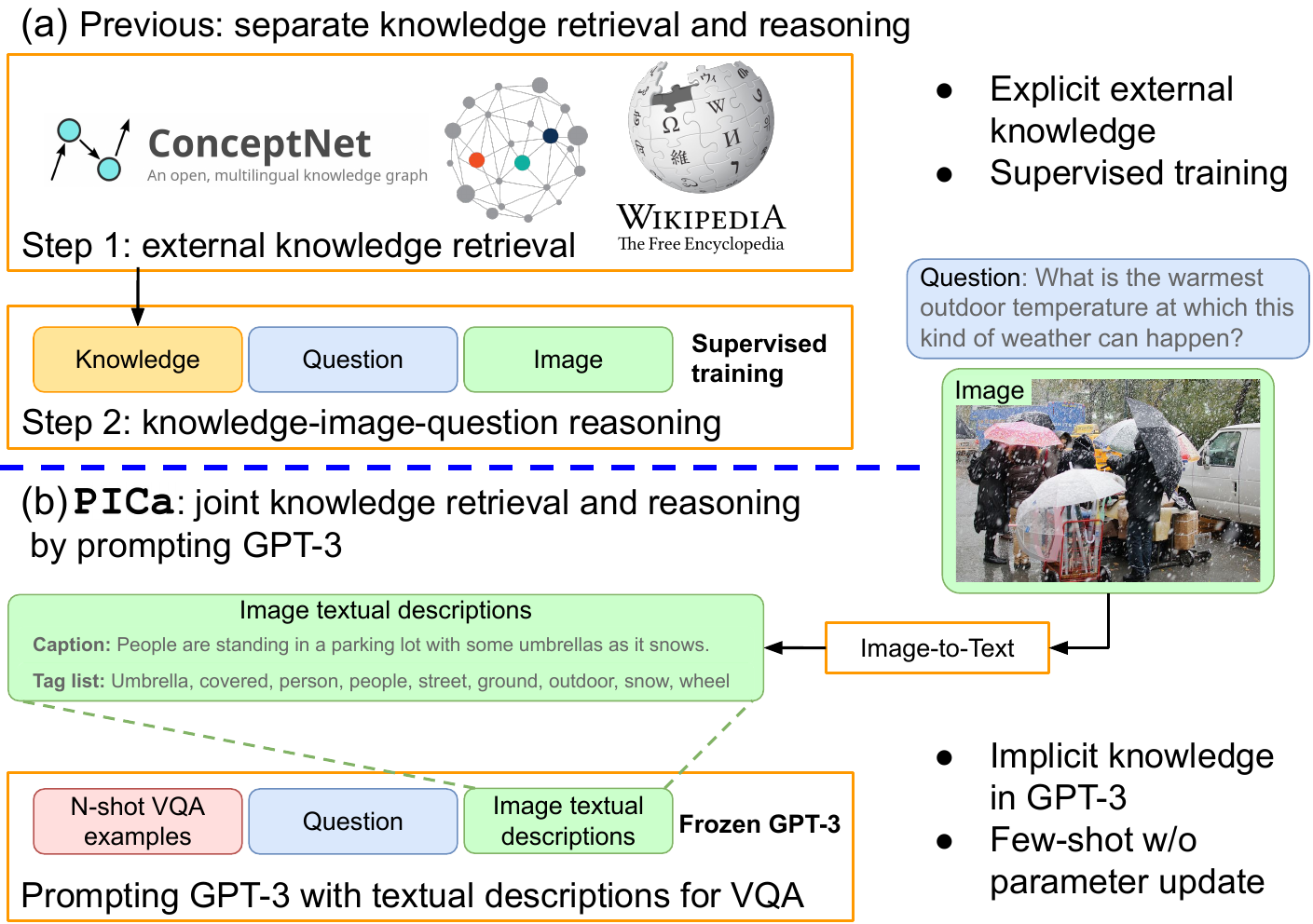}}
\end{center}
\vspace{-0.15in}
\caption{\small Comparison between previous methods and our proposed \oursmodel~for knowledge-based VQA. (a) Previous methods adopt a two-step approach, which first retrieves the external knowledge, then reasons over the selected knowledge, the input image, and question for answer prediction. (b) Alternatively,~\oursmodel~directly prompts GPT-3 to jointly acquire and reason over the relevant knowledge. We convert images into textual descriptions that GPT-3 can understand, and adapt GPT-3 to solve the task by providing only a few in-context VQA examples during inference time.	}
\vspace{-0.15in}
\label{fig:intro}
\end{figure}

The problem of knowledge-based visual question answering (VQA)~\cite{marino2019ok} extends the standard VQA task~\cite{VQA_15} by asking questions that require outside knowledge beyond the image content to answer. To obtain such knowledge, existing methods~\cite{zhu2020mucko,garderes2020conceptbert,marino2021krisp,wu2021multi} first retrieve external knowledge from multiple resources, such as Wikipedia articles and ConceptNet concepts~\cite{speer2017conceptnet}. Based on this, joint reasoning over the retrieved knowledge and the image-question pair is performed to predict the answer, as shown in Figure~\ref{fig:intro}(a).

However, this two-step approach could be sub-optimal. For example, the image-question feature used for knowledge retrieval in the first step may not match their representations in the second reasoning step, which leads to noisy or even irrelevant retrieved knowledge. The re-embedded textual feature of the retrieved knowledge might also deviate from its original meaning in the knowledge source. Such mismatches potentially limit the VQA performance. Furthermore, learning a good joint knowledge-image-question representation requires sufficient training data, thus making it difficult to transfer to new types of questions. In this study, we explore an alternative approach inspired by the intriguing properties of recent language models. Specifically, large-scale language models such as GPT-3~\cite{brown2020language} have shown powerful abilities in NLP tasks, such as knowledge retrieval~\cite{wang2020language} and question answering~\cite{brown2020language}. More impressively, they are also strong few-shot learners, \ie, the model can quickly adapt to new tasks by using only a few in-context examples. 

Inspired by this, we propose~\oursmodel,\footnote{\textbf{\texttt{P}}rompting GPT-3 via the use of \textbf{\texttt{I}}mage \textbf{\texttt{Ca}}ptions}
a simple yet effective method that unifies the above knowledge retrieval and reasoning steps with the help of GPT-3. Instead of using \emph{explicit} and \emph{structured} knowledge bases (KBs) as in previous work,~\oursmodel~treats GPT-3 as an \emph{implicit} and \emph{unstructured} KB~\cite{petroni2019language} via prompt engineering. 
Specifically, we convert images into textual descriptions (\emph{i.e.}, captions or tags) that GPT-3 can understand, and then query GPT-3 to directly predict the answer based on the question and textual descriptions, as shown in Figure~\ref{fig:intro}(b). 
Instead of supervised fine-tuning,~\oursmodel~inherits the \emph{few-shot} learning ability from GPT-3, and adapts to the VQA task with only a few in-context examples during inference time. Empirically, we show that GPT-3 can implicitly retrieve relevant knowledge, and effectively reason over the question and context for answer prediction. To further boost performance, we have carefully investigated: ($i$) how image contexts can be effectively represented as textual descriptions, and ($ii$) how to better select in-context examples and use multi-query ensemble to further unleash the power of GPT-3. 

We conduct comprehensive experiments on the OK-VQA dataset~\cite{marino2019ok}. With a pre-trained captioning model (VinVL)~\cite{zhang2021vinvl},~\oursmodel~achieves an accuracy of $46.9\%$ in a few-shot manner, an absolute improvement of $7.5$ points when compared with supervised state of the art~\cite{wu2021multi}. When enhanced with predicted image tags, the performance can be further boosted to $48.0$. We also provide detailed ablation study and qualitative analysis to understand the effectiveness of \oursmodel.

Our main contributions are summarized as follows.
($i$) We present~\oursmodel, a simple yet effective method to use GPT-3 for knowledge-based VQA, demonstrating the first use of GPT-3 for multimodal tasks. 
($ii$) \oursmodel~represents images as textual descriptions, and enhances the performance of GPT-3 via in-context example selection and multi-query ensemble.
($iii$) \oursmodel~achieves $48.0\%$ accuracy on OK-VQA in a \emph{few-shot} manner, lifting up the state of the art of $39.4\%$ by a significant margin. It also achieves a decent few-shot performance on VQAv2~\cite{goyal2017making}.
\section{Related Work}
\noindent\textbf{Knowledge-based VQA.}
Knowledge-based VQA requires external knowledge in addition to the image content to answer a question. Early explorations include KB-VQA~\cite{wang2015explicit} and F-VQA~\cite{wang2017fvqa}. The more recent OK-VQA dataset~\cite{marino2019ok} is built on COCO images~\cite{lin2014microsoft}, and the input questions cover a wide range of knowledge categories. Previous studies~\cite{wang2015explicit,narasimhan2018straight,wang2017fvqa,narasimhan2018out,zhu2020mucko,li2020boosting,marino2021krisp,wu2021multi} proposed various ways of retrieving and using the knowledge, and considered it necessary to use multiple knowledge resources, such as Wikipedia, ConceptNet~\cite{speer2017conceptnet}, Google images, and the implicit knowledge from language models~\cite{devlin2018bert,zhu2015aligning}, to cover the relevant knowledge in questions. After external knowledge retrieval, studies have focused on reasoning over the knowledge acquired and the input image-question pair for answer prediction, where graph convolution network has been shown to be an effective way for multimodal fusion~\cite{narasimhan2018out,zhu2020mucko}. More recently, KRISP~\cite{marino2021krisp} proposed to retrieve the implicit knowledge stored in pre-trained language models as a supplementary knowledge resource to the structured knowledge base. MAVEx~\cite{wu2021multi} presented an answer validation approach to make better use of the noisy retrieved knowledge. The above two-step approaches may not get the most relevant knowledge in the retrieval step, and fail to best encode the knowledge for QA in the reasoning step. In this study, we combine the two steps, and present a model that jointly acquires and processes the knowledge for VQA by prompting GPT-3.

\vspace{2mm}
\noindent\textbf{Multimodal few-shot learning.}
GPT-3~\cite{brown2020language} has shown astonishing in-context few-shot learning capabilities. Recently, Frozen~\cite{tsimpoukelli2021multimodal} is proposed to extend such few-shot abilities to vision-and-language tasks by reusing a pre-trained language model. Specifically, Frozen starts with a GPT-like language model with 7 billion parameters pre-trained on a large text corpus. Then, Frozen freezes the language model, and trains a visual encoder to project input images to visual features that the language model can understand. The visual encoder is trained with the image captioning task~\cite{sharma2018conceptual}, with gradients being back-propagated from the frozen language model. Frozen presents the first ever multimodal few-shot learner, and performs much better than random guess on tasks such as VQA. Despite interesting observations, the performance is far from satisfactory compared to the state of the art. For example, Frozen only achieves an accuracy of $12.6\%$ on the OK-VQA dataset~\cite{marino2019ok}. 
Our idea of utilizing the pre-trained language model is closely related to Frozen. However, we push the limit, and investigate a much stronger language model, with a focus on the knowledge-based VQA task. To this end, we present the first few-shot approach that surpasses the supervised state of the art.
\section{Approach}
\begin{figure*}[t]
\begin{center}
  \centerline{\includegraphics[width=16cm]{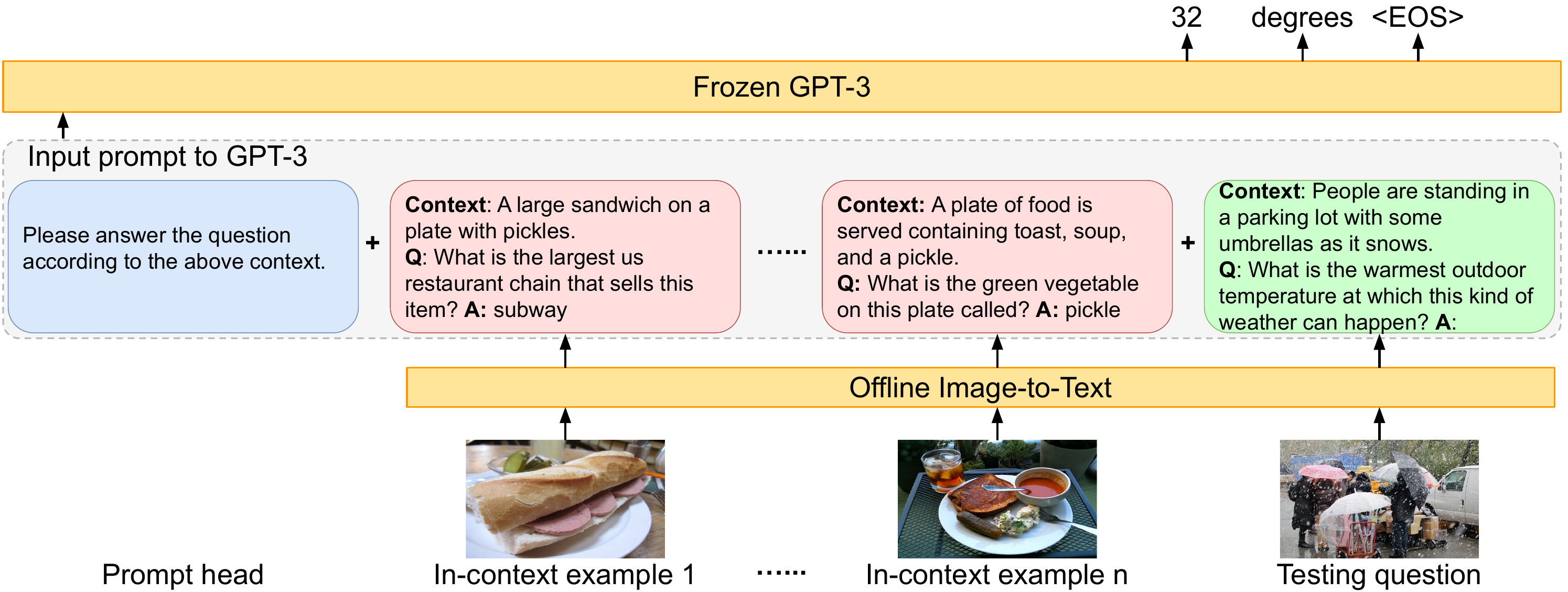}}
\end{center}
\vspace{-0.25in}
    \caption{Inference-time interface of~\oursmodel~for $n$-shot VQA. The input prompt to GPT-3 consists of a prompt head $\hv$ ({\color{blue}blue} box), $n$ in-context examples ($\{\xv_i,\yv_i\}_{i=1}^n$) ({\color{red}red} boxes), and the VQA input $\xv$ ({\color{green}green} box). The answer $\yv$ is produced in an open-ended text generation manner. \oursmodel~supports zero-/few-shot VQA by including different numbers of in-context examples in prompt.}
\vspace{-0.1in}
\label{fig:prompt}
\end{figure*}

\subsection{GPT-3 for In-context Learning}
GPT-3~\cite{brown2020language} has shown powerful in-context few-shot learning abilities. Instead of fine-tuning a pre-trained model to adapt it to a downstream task, in-context few-shot learners quickly adapt to new tasks with just a few examples at inference time, and require no parameter updates. Concretely, during inference, the target of the new task $\yv$ is directly predicted conditioned on the given context $\Ccal$ and the new task's input $\xv$, as a text sequence generation task. Note that all the $\Ccal, \xv, \yv$ are text sequences. For example, $\yv=(y^1,\cdots,y^T)$. Therefore, at each decoding step $t$,
\begin{equation*}
    y^t = \arg\max_{y^t}p_{\text{LM}}(y^t|\Ccal,\xv,y^{<t})\,,
\end{equation*}
where LM represents the weights of the pre-trained language model, which are frozen for all new tasks. The context $\Ccal=\left\{\hv,\xv_1,\yv_1,\cdots,\xv_n,\yv_n \right\}$ consists of an optional prompt head $\hv$ and $n$ in-context examples ($\{\xv_i,\yv_i\}_{i=1}^n$) from the new task. Inspired by GPT-3's strong few-shot ability and the diverse knowledge it contains, we present below the use of GPT-3 for few-shot knowledge-based VQA in detail.

\vspace{-1mm}
\subsection{GPT-3 for VQA}
\vspace{-1mm}
One challenge with using GPT-3 for VQA is that GPT-3 does not inherently understand image input. Empirically, we show that converting image context into textual descriptions leads to a strong baseline for VQA. Figure~\ref{fig:prompt} shows the inference-time interface of~\oursmodel, which approaches the VQA task by prompting GPT-3 with a constructed input prompt. The prompt is a word sequence that consists of context $\Ccal$ (with a prompt head $\hv$ and $n$ in-context examples $\{\xv_i,\yv_i\}_{i=1}^n$) and VQA input $\xv$. Specifically, we first adopt state-of-the-art captioning (or tagging) models to translate the VQA image into captions (or a list of tags). As shown in the {\color{green}green} box, the VQA input $\xv$ is the concatenation of the translated image context string (``\texttt{Context: People are standing in a parking lot with some umbrellas as it snows.}'') and the question string (``\texttt{Q: What is the warmest temperature at which this weather can happen? A:}''). The target $\yv$ is the output answer (``\texttt{32 degrees}''). The answer is produced in an open-ended text generation manner, \ie, the answer could contain an arbitrary number of words selected from the entire vocabulary of GPT-3. The context $\Ccal$ starts with a prompt head $\hv$, which is a fixed string (``\texttt{Please answer the question according to the above context.}'') for all samples, as shown in the {\color{blue}blue} box. The remaining part of $\Ccal$ is the concatenation of $n$ in-context example strings $(\{\xv_i,\yv_i\}_{i=1}^n)$ like in the {\color{red}red} boxes. We then concatenate $\Ccal$ with the VQA input $\xv$ shown in the {\color{green}green} box to generate the prompt. GPT-3 takes the constructed prompt text as input, implicitly retrieving and reasoning the knowledge from the language model, and predicts the answer $\yv$ as an open-ended text generation task.

\begin{table*}[t]
\small
\centering
\vspace{-0.0in}
\begin{tabular}{ l l l c || c }
    \hline
    Method & Image Repr. & Knowledge Resources & Few-shot & Accuracy  \\
    \hline
    MUTAN+AN~\cite{ben2017mutan} & Feature Emb. & Wikipedia & \xmark & 27.8 \\
    Mucko~\cite{zhu2020mucko} & Feature Emb. & Dense Captions & \xmark & 29.2 \\
    ConceptBert~\cite{garderes2020conceptbert} & Feature Emb. & ConceptNet & \xmark & 33.7 \\
    ViLBERT~\cite{lu2019vilbert} & Feature Emb. & None & \xmark & 35.2 \\
    KRISP~\cite{marino2021krisp} & Feature Emb. & Wikipedia + ConceptNet & \xmark & 38.9 \\
    MAVEx~\cite{wu2021multi} & Feature Emb. & \small{Wikipedia + ConceptNet + Google Images} & \xmark & \underline{39.4} \\
    \hline
    Frozen~\cite{tsimpoukelli2021multimodal} & Feature Emb. & Language Model (7B) & \cmark & 12.6 \\
    \textbf{\texttt{PICa-Base}} & Caption & GPT-3 (175B) & \cmark & 42.0  \\
    \textbf{\texttt{PICa-Base}} & Caption+Tags & GPT-3  (175B)& \cmark & 43.3 \\
    \textbf{\texttt{PICa-Full}} & Caption & GPT-3 (175B) & \cmark & {46.9} \\
    \textbf{\texttt{PICa-Full}} & Caption+Tags & GPT-3 (175B) & \cmark & \textbf{48.0} \\
    \hline
\end{tabular}
\vspace{-2mm}
\caption{\small Results on the OK-VQA test set~\cite{marino2019ok}. The upper part shows the supervised state of the art, and the bottom part includes the few-shot performance of Frozen~\cite{tsimpoukelli2021multimodal} and the proposed \oursmodel~method.}
\vspace{-0.0in}
\label{table:okvqa}
\end{table*}

\begin{table*}[t]
\small
\centering
\vspace{-0.0in}
\begin{tabular}{ c l l || c c c c c | c }
    \hline
    & Method & Image Repr. & $n$=0 & $n$=1 & $n$=4 & $n$=8 & $n$=16 & Example engineering \\
    \hline
    (a) & Frozen~\cite{tsimpoukelli2021multimodal} & Feature Emb. & 5.9 & 9.7 & 12.6 & - & - & \xmark \\
    (b) & \textbf{\texttt{PICa-Base}} & Caption & 17.5 & 32.4 & 37.6 & 39.6 & 42.0 &  \xmark \\
    (c) & \textbf{\texttt{PICa-Base}} & Caption+Tags & 16.4 & 34.0 & 39.7 & 41.8 & 43.3 &  \xmark \\
    \hline
    (d) & \textbf{\texttt{PICa-Full}} & Caption & 17.7 & 40.3 & 44.8 & 46.1 & 46.9 & \cmark \\
    (e) & \textbf{\texttt{PICa-Full}} & Caption+Tags & 17.1 & 40.8 & 45.4 & 46.8 & \textbf{48.0} & \cmark \\
    \hline
\end{tabular}
\vspace{-2mm}
\caption{\small The few-shot in-context learning results on the OK-VQA dataset. The ``Example engineering'' column indicates whether the method needs the access to an in-context example pool that contains more than $n$ in-context examples from the new task.}
\vspace{-3mm}
\label{table:shots}
\end{table*}

\subsection{In-context Examples}
Empirically, feeding more in-context examples leads to better few-shot performance~\cite{brown2020language,tsimpoukelli2021multimodal}. However, the number of available examples in the new task and the model's max input length jointly constrain the max number of examples $n$ in the prompt. In practice, we observe that the max input length more often limits the max $n$ we can take, \ie, there are usually more available examples than the ones that a language model can take (\eg, $n=16$). To better use these available examples, below, we explore two approaches: ($i$) improving the example quality by careful in-context example selection~\cite{liu2021makes}, and ($ii$) using more examples via multi-query ensemble.

\vspace{2mm}
\noindent\textbf{In-context example selection.}
In-context example selection~\cite{liu2021makes} tries to search for the best examples for each inference-time input $\xv$ among all available examples. We consider an in-context example $\xv_i$ a good one if it has a similar question feature as $\xv$. Specifically, we leverage the CLIP model (ViT-B/16 variant)~\cite{radford2021learning} for similarity calculation. Given an inference-time question, we obtain its textual feature with the text encoder of CLIP~\cite{radford2021learning}, and compute its cosine similarity with the questions in all available in-context examples. We then average the question text similarity with the image visual similarity to guide the example selection. We select the top $n$ questions with the highest similarities, and use the corresponding examples as the in-context examples.

\vspace{2mm}
\noindent\textbf{Multi-query ensemble.}
An alternative approach to better use available examples is multi-query ensemble. Given an inference-time example $\xv$, we use $n*k$ in-context examples to generate $k$ prompts. By prompting GPT-3 for $k$ times, we obtain $k$ answer predictions instead of $1$, where $k$ is the number of queries to ensemble. Among the $k$ answer predictions, we select the one with the highest sum of log-probability $\sum_t \log p_{\text{LM}}(y^t)$ as the final answer~\cite{chen2021evaluating}. The multi-query ensemble can be seamlessly used together with the in-context example selection. By selecting the top $n*k$ examples and distributing them into $k$ prompts, we combine the two methods and obtain the gains from both approaches. 
\section{Experiments on OK-VQA}

\subsection{Dataset and Setup}
\noindent\textbf{Dataset.}
OK-VQA~\cite{marino2019ok} is currently the largest knowledge-based VQA dataset, with 14,055 image-question pairs. Questions are manually filtered to ensure that outside knowledge is required to answer the question. Each question has $5$ ground-truth answers. The soft accuracy from VQAv2~\cite{goyal2017making} is used for  evaluation.

\vspace{2mm}
\noindent\textbf{Setup.}
We compare two variants of our method.
\vspace{-2pt}
\begin{itemize} 
\setlength\itemsep{-2pt}
\item \textbf{\texttt{PICa-Base}}. This method uses prompts shown in Figure~\ref{fig:prompt}. We represent images either as captions with VinVL~\cite{zhang2021vinvl}, or enhance captions with tags predicted by the public Microsoft Azure tagging API\footnote{Public Azure Tagging \& Captioning API: \url{https://westus.dev.cognitive.microsoft.com/docs/services/computer-vision-v3-2}}. 
In-context examples are randomly selected. \item \textbf{\texttt{PICa-Full}}. This is the full model that includes in-context example selection and multi-query ensemble.
\vspace{-2pt}
\end{itemize}

\subsection{Comparison with State-of-the-art}
Table~\ref{table:okvqa} summarizes the results on the OK-VQA dataset. The upper part of the table contains models that are trained on the complete OK-VQA training set in a supervised manner. The lower part lists the few-shot results. The column ``Image Repr.'' indicates how the image is represented for VQA. ``Feature Emb.'' refers to the conventional approach that encodes the image as feature vectors with a trainable network. Due to the high cost of end-to-end fine-tuning GPT-3, we convert images into text sequences that GPT-3 can understand. ``Caption'' means the caption generated by the VinVL-base model fine-tuned on the COCO-train14 split. ``Tags'' indicates the tags predicted by the tagging API. The column ``Knowledge Resource'' includes the external knowledge resources used. Most previous methods 
involve explicit knowledge retrieval from external knowledge resources, \eg, ``Wikipedia'' and ``ConceptNet.'' Alternatively, few-shot methods directly use pre-trained language models to acquire and process the knowledge. We summarize our observations as follows. 
\vspace{-2pt}
\begin{itemize} 
\setlength\itemsep{-2pt}
\item Our method surpasses the state of the art~\cite{wu2021multi} with no model fine-tuning. 
\textbf{\texttt{PICa-Full}} achieves an accuracy of $48.0\%$ with 16 in-context VQA examples, compared with the supervised state-of-the-art accuracy of $39.4\%$ trained on the entire OK-VQA training set. The superior performance shows the power of \emph{implicit} knowledge retrieval and reasoning with GPT-3, in contrast to retrieving the external knowledge \emph{explicitly}.
\item Compared with \textbf{\texttt{PICa-Base}} that uses randomly selected in-context examples, \textbf{\texttt{PICa-Full}} achieves better performance by more effectively using the available in-context examples. \textbf{\texttt{PICa-Full}} improves over \textbf{\texttt{PICa-Base}} from $42.0\%$ to $46.9\%$ with captions, and from $43.3\%$ to $48.0\%$ with both captions and tags. Detailed ablation studies are provided in Table~\ref{table:ensemble}.
\vspace{-2pt}
\end{itemize}


\subsection{Few-shot Ability}
In this section, we zoom in the lower part of Table~\ref{table:okvqa}, and analyze the model's few-shot abilities on the OK-VQA dataset in Table~\ref{table:shots}. 
The upper part of the table contains results with in-context examples randomly selected from an example pool, \ie, the \emph{strict} few-shot setting. In practice, we often have more than $n$ examples at hand, and selecting better in-context examples leads to better few-shot performance. The lower part of the table shows the results of the methods with in-context example selection and multi-query ensemble. $n$ is the number of in-context examples, and is also known as ``the number of shots''. We experiment with $n$ ranging from $0$ to $16$. $n=16$ is roughly the max number of examples that GPT-3 can take, with a max input length of $2049$. We re-select in-context examples of shorter lengths if any prompt exceeds the max input length limit, which rarely happens with $n=16$.

We observe that more shots generally lead to better performance, \eg, $40.8\%$ when $n=1$ to $48.0\%$ when $n=16$ in row (e). This observation supports our motivation of using more examples whenever possible. Compared with \textbf{\texttt{PICa-Base}}, \textbf{\texttt{PICa-Full}} yields consistent $5\%$ accuracy improvements across all cases.
\begin{table}[t]
\small
\centering
\vspace{-0.0in}
\begin{tabular}{ c l || c c}
    \hline
    & Image Repr. & \textbf{\texttt{Base}} & \textbf{\texttt{Full}} \\
    \hline
    (a) & Blind & 24.2 & 30.1 \\
    (b) & Tags & 39.3 & 44.6 \\
    (c) & VinVL-Caption-CC & 37.0 & 44.0 \\
    (d) & API-Caption & 39.1 & 45.2\\
    (e) & VinVL-Caption-COCO & 42.0 & 46.9 \\
    (f) & GT-Caption-1$^\dagger$ & 42.1 & 48.7\\
    (g) & GT-Caption-5$^\dagger$ & 48.0 & \underline{53.3} \\
    \hline
    (h) & VinVL-Caption-CC+Tags & 41.5 & 46.0 \\
    (i) & API-Caption+Tags & 41.9 & 47.4 \\
    (j) & VinVL-Caption-COCO+Tags & 43.3 & \textbf{48.0}\\
    \hline
\end{tabular}
\vspace{-2mm}
\caption{\small Ablation study on different textual representations for images on OK-VQA. ($\dagger$) indicates the oracle performance. The best accuracy and oracle performance are highlighted in bold and underline, respectively.}
\vspace{-2mm}
\label{table:content}
\end{table}

\subsection{Textual Representation for Images}
We provide ablation study on how to best represent images in the textual format for GPT-3. Specifically, we compare the following methods.
\vspace{-2pt}
\begin{itemize} 
\setlength\itemsep{-2pt}
\item \textbf{Blind.} Blind is the baseline that uses an empty string to represent the image, which indicates the VQA performance without looking at the image. We also use the question similarity alone for in-context example selection to enforce the blind setting.
\item \textbf{Tags.} We represent the image as a list of tags predicted by an automatic tagging model. All tags are concatenated as a string with a comma separator.
\item \textbf{VinVL-Caption-COCO.} This is the caption used for the results in Tables~\ref{table:okvqa} and \ref{table:shots}. We fine-tune the VinVL-base pre-trained checkpoint with the COCO 2014 training set to obtain the image captions on the OK-VQA test set, which contains images from COCO 2014 validation set.
\item \textbf{VinVL-Caption-CC.} To follow a more \emph{strict} few-shot setting and avoid seeing images from the same COCO dataset, we train a VinVL-base captioning model with the Conceptual Captions dataset~\cite{sharma2018conceptual}.
\item \textbf{API-Caption.} This indicates the caption generated from the public Azure API in Footnote 2.
\item \textbf{GT-Caption.} We include the ground-truth COCO captions as the oracle with ideal image descriptions. We use either $1$ randomly sampled ground truth or the concatenation of all $5$ captions as the image description.
\item \textbf{Caption+Tags.} We represent the image as the concatenation of the caption string and tag list string.
\vspace{-2pt}
\end{itemize}
Table~\ref{table:content} shows the OK-VQA accuracies with different formats of textual representations. 
We summarize our findings as follows.
($i$) All formats of textual descriptions in rows (b-j) provide a good image representation, as all methods significantly outperform the blind baseline in row (a), which only takes the question-answer pair as input.
($ii$) Despite never seeing COCO images, the VinVL-Caption-CC method in row (c) achieves a decent accuracy of $37.0\%$ with $n=16$. The performance can be further improved to $44.0\%$ when including in-context example selection and multi-query ensemble, which surpasses the supervised state of the art.
($iii$) When comparing different predicted captions being used, VinVL-Caption-COCO in row (e) achieves the best performance. In general, we find that detailed and thorough descriptions lead to better VQA performance.
($iv$) The ground-truth COCO caption (row (f)) provides a more accurate description of the image, thus leading to an oracle accuracy of $48.7\%$. Concatenating all the ground-truth captions as shown in row (g) provides a more thorough description of the image content, thus further improving the accuracy to $53.3\%$.
($v$) Inspired by the effectiveness of concatenating multiple captions, we also experiment with combining multiple formats of textual descriptions, as shown in the bottom part of the table. We observe that combining captions with tags provides complementary information and helps VQA. For example, in row (j), combining VinVL captions and tags results in a $16$-shot accuracy of $48.0\%$, compared with the accuracy in rows (b,e) of $44.6\%$ and $46.9\%$, respectively. Similar improvements are observed in other combinations as shown in rows (h,i).

\begin{table}[t]
\small
\centering
\begin{tabular}{ c l || c c }
    \hline
    & Methods & CLIP & RoBERTa \\
    \hline
    (a) & Random & \multicolumn{2}{c}{43.3}  \\
    (b) & Question & 45.8 & 45.4 \\
    (c) & Question$_\text{Dissimilar}$ & 40.1 & 40.9 \\
    (d) & Question+Answer$^\dagger$ & \underline{49.1} & 48.4 \\
    \hline
    (e) & Image only & 44.1 & - \\
    (f) & Image+Question & \textbf{46.5} & - \\
    \hline\end{tabular}
\vspace{-2mm}
\caption{\small Ablation study on different in-context example selection methods on OK-VQA with $16$ examples. ($\dagger$) indicates the oracle performance. The best accuracy and oracle performance are highlighted in bold and underline, respectively.}
\label{table:qaselect}
\vspace{-2mm}
\end{table}

\begin{figure*}[t]
\begin{center}
  \centerline{\includegraphics[width=16cm]{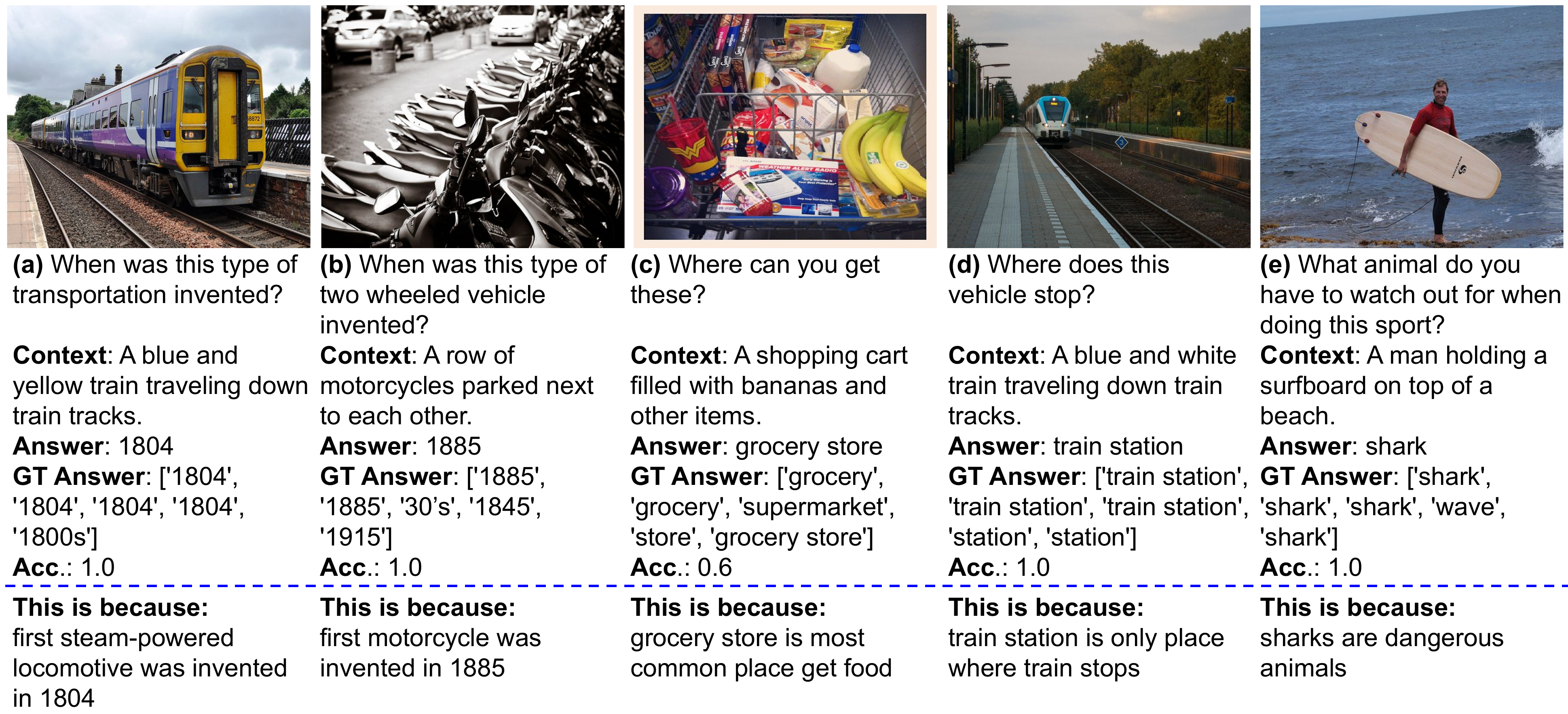}}
\end{center}
\vspace{-0.3in}
    \caption{\small Qualitative examples of our proposed \textbf{\texttt{PICa}} method on the OK-VQA dataset. The upper part shows GPT-3 predicted answers, and the bottom part includes the answer rationales generated in a zero-shot manner.
	}
\vspace{-0.1in}
\label{fig:okvisu}
\end{figure*}

\subsection{Example Selection and Multi-query Ensemble}
\noindent\textbf{In-context example selection.}
Results are summarized in Table~\ref{table:qaselect}. 
Row (a) shows the \textbf{\texttt{PICa-Base}} performance where examples are randomly selected. Row (b) selects examples based on the similarity of the question textual features. We experiment with choosing the most dissimilar examples in row (c), and observe that ``bad'' examples indeed lead to worse performance. Row (d) shows an oracle number that includes the answer similarity in example selection. This serves as an upper bound, and shows that properly selecting examples can significantly improve the VQA accuracy. Rows (e,f) include image visual features for example selection. Specifically, row (e) selects examples based on image feature similarity computed by CLIP~\cite{radford2021learning}. Row (f) presents the approach in \textbf{\texttt{PICa-Full}} that jointly considers the question and image similarities. 

The improvement of row (b) over row (a) shows that in-context example selection indeed helps few-shot VQA. Row (c) presents an expected low accuracy of $40.1\%$ with dissimilar examples, indicating the effectiveness of using question similarity to guide the in-context example selection. By selecting the ``ideal'' examples in row (d), the oracle accuracy reaches $49.1\%$. We observe a slightly better performance when computing the feature with the CLIP text encoder~\cite{radford2021learning} than a pure language model RoBERTa~\cite{liu2019roberta}. 
Example selection with image similarity alone also improves the random baseline, as shown in row (e). The improvement is smaller than using question similarity alone in row (b), as question similarity is more informative in the VQA task. \textbf{\texttt{PICa-Full}} jointly considers the question and image similarities, and further improves the accuracy to $46.5\%$, as in row (f).

\vspace{2mm}
\noindent\textbf{Multi-query ensemble.}
Multi-query ensemble allows the model to use more in-context examples at inference time, thus potentially further improving the performance. Multi-query ensemble can be seamlessly used together with in-context example selection. Table~\ref{table:ensemble} shows the results of combining them together. Rows (a,b) are the baseline results without multi-query ensemble. Rows (c-d) show that by increasing the number of prompts $k$, the OK-VQA accuracy can be consistently improved on all shot numbers $n$.

\begin{figure*}[t]
\begin{center}
  \centerline{\includegraphics[width=16cm]{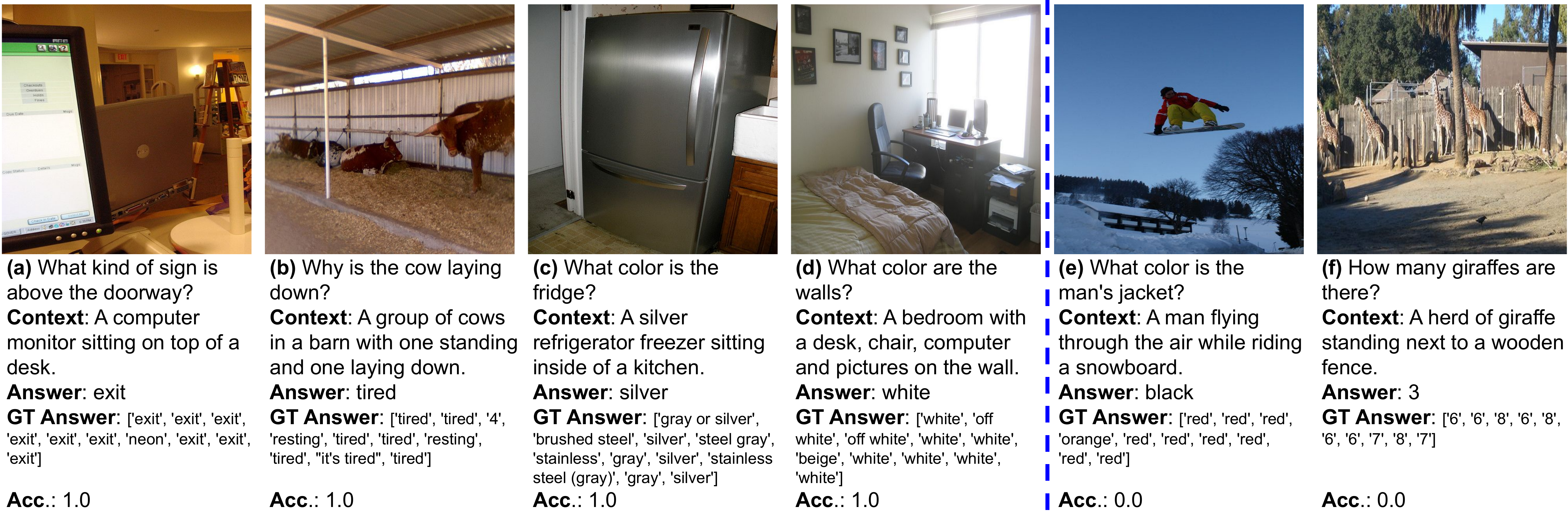}}
\end{center}
\vspace{-0.3in}
    \caption{\small Representative success~(the left four examples) and failure (the right two examples) cases of \textbf{\texttt{PICa}} on the VQAv2 dataset.
	}
\vspace{-0.1in}
\label{fig:vqavisu}
\end{figure*}

\subsection{Qualitative Analysis}
\label{sec:example}

\noindent\textbf{Representative cases.}
The upper part of Figure~\ref{fig:okvisu} shows some qualitative examples of our \textbf{\texttt{PICa}} predictions. We observe that \textbf{\texttt{PICa}} works well on questions that require different kinds of external knowledge. For example, in Figure~\ref{fig:okvisu}(a), GPT-3 understands that ``\texttt{this type of transportation}'' in the question refers to the ``\texttt{train}'' in the image, and provides the correct answer that ``\texttt{train was invented in 1804}''.\footnote{\url{www.google.com/search?q=when+was+train+invented}} Similarly, in Figure~\ref{fig:okvisu}(b), the model knows that ``\texttt{motorcycle was invented in 1885}''.\footnote{\url{www.google.com/search?q=when+was+motorcycle+invented}} Other than factual or encyclopedia knowledge, \textbf{\texttt{PICa}} also works well on questions that need commonsense knowledge. For example, in Figure~\ref{fig:okvisu}(c), the model understands that people can get bananas from grocery stores. The disagreement among ground-truth answers in this example also shows that the open-ended answer generation could produce different formats of the correct answer, making the evaluation challenging. Similarly, in Figures~\ref{fig:okvisu}(d,e), the model correctly answers the question with the implicit knowledge of ``\texttt{train stops at the train station}'' and ``\texttt{there could be sharks in the sea when surfacing}''.

\begin{table}[t]
\small
\centering
\centering
\begin{tabular}{ c l || c c c }
    \hline
    & \# of ensembles & $n$=1 & $n$=4 & $n$=16\\
    \hline
    (a) & $k$=1 w/o selection & 34.0 & 39.7 & 43.3  \\
    (b) & $k$=1 & 36.4 & 43.0 & 46.5  \\
    (c) & $k$=3 & 40.0 & 45.2 & 47.7  \\
    (d) & $k$=5 & 40.8 & 45.4 & \textbf{48.0}  \\
    \hline\end{tabular}
\vspace{-2mm}
\caption{\small The multi-query ensemble performance on OK-VQA. Experiments are based on performing in-context example selection with question and image similarity. $k$ is the number of prompts to ensemble. Row (a) corresponds to \textbf{\texttt{PICa-Base}}, while row (d) correspond to \textbf{\texttt{PICa-Full}}.}
\label{table:ensemble}
\vspace{-3mm}
\end{table}

\vspace{2mm}
\noindent\textbf{Answer rationale.}
One may wonder how GPT-3 correctly answers the knowledge-based questions in an open-ended manner without being fine-tuned for the task. The inaccessibility of the GPT-3 raw model makes it difficult to conduct an in-depth analysis of the language model's behavior. Alternatively, we perform answer rationale prediction~\cite{li2018vqa,park2018multimodal,zellers2019recognition} in a zero-shot manner, and generate answer rationale as an open-ended text generation task. Specifically, we construct a prompt that is the concatenation of question string $\xv$, predicted answer $\yv$, and a prompt head ``\texttt{This is because}''. We then take GPT-3's generated text as the answer rationale.

The bottom part of Figure~\ref{fig:okvisu} shows the rationales for the predicted answers. We observe that GPT-3 generates reasonable rationales for questions that need different types of knowledge. For example, in Figure~\ref{fig:okvisu}(a), the rationale is the core encyclopedia knowledge that ``\texttt{the first locomotive was invented in 1804}''. Figure~\ref{fig:okvisu}(c) shows an example that the model provides the commonsense knowledge of ``\texttt{grocery store is a common place to get food}''.

\section{Experiments on VQAv2}
Despite the good performance on knowledge-based VQA, one limitation of our method is that the image is abstracted as text. Captions or tags only provide a partial description of the image, and might miss important visual details necessary for question answering, such as questions on detailed visual attribute prediction. In this section, we benchmark~\oursmodel~on the VQAv2 dataset~\cite{goyal2017making} that contains questions focusing on the detailed image contents.
\begin{table}[t]
\small
\centering
\vspace{-0.0in}
\begin{tabular}{ l l c || c }
    \hline
    Method & Image Repr. & Few-shot & Acc. \\
    \hline
    \small{Oscar~\cite{li2020oscar}} & Feature Emb. & \xmark & 73.8 \\
    \hline
    Frozen & Feature Emb. & \cmark & 38.2 \\
    \textbf{\texttt{PICa-Base}} & Caption & \cmark & 53.2 \\
    \textbf{\texttt{PICa-Base}} & Caption+Tags & \cmark & 54.3 \\
    \textbf{\texttt{PICa-Full}} & Caption & \cmark & 55.9 \\
    \textbf{\texttt{PICa-Full}} & Caption+Tags & \cmark & \textbf{56.1} \\
    \textbf{\texttt{PICa-Full}}$^\dagger$ & GT-Caption-5 & \cmark & \underline{59.7} \\
    \hline
\end{tabular}
\vspace{-2mm}
\caption{\small Results on the VQAv2 validation set. The upper part shows the supervised state of the art. The bottom part shows the few-shot accuracy. ($\dagger$) indicates the oracle performance.}
\vspace{-2mm}
\label{table:vqav2}
\end{table}
\vspace{2mm}
\noindent\textbf{Dataset and setup.} 
The VQAv2 dataset~\cite{goyal2017making} annotates question-answer pairs based on the COCO image corpus~\cite{lin2014microsoft}. VQAv2 questions are designed to be highly relevant to the image content. It reports the human performance of $40.8\%$ with questions only, and $57.5\%$ with questions and captions, compared with $83.3\%$ with both questions and images. We follow Frozen~\cite{tsimpoukelli2021multimodal}, and report the accuracy on the validation set. 
Instead of treating VQA as a classification task over a pre-selected answer vocabulary~\cite{goyal2017making,li2020oscar}, we predict the answer in an open-ended text generation manner.

\vspace{2mm}
\noindent\textbf{Results.}
Table~\ref{table:vqav2} summarizes our results on VQAv2. \textbf{\texttt{PICa-Full}} achieves an accuracy of $56.1\%$, surpassing the previous few-shot accuracy of $38.2\%$ by a significant margin~\cite{tsimpoukelli2021multimodal}. The proposed in-context example selection and multi-query ensemble methods also work well on the VQAv2 dataset (\cf, \textbf{\texttt{PICa-Base}}: $54.3\%$, \textbf{\texttt{PICa-Full}}: $56.1\%$). Compared with the supervised performance of $73.8\%$ by Oscar~\cite{li2020oscar}, the proposed method is still around $17\%$ lower in accuracy with failure cases discussed in Figure~\ref{fig:vqavisu}. Nonetheless, the promising few-shot results show that the proposed approach is one strong baseline in approaching few-shot vision-language tasks.

\vspace{2mm}
\noindent\textbf{Limitations.}
Figures~\ref{fig:vqavisu}(a-d) and (e,f) show qualitative examples of the success and failure cases of \textbf{\texttt{PICa-Full}}, respectively. 
A subset of VQAv2 questions can be answered with commonsense knowledge, where~\oursmodel~generally performs well. For example, the implicit knowledge of ``\texttt{the sign above doorway can be the exit sign}'' in Figure~\ref{fig:vqavisu}(a) and ``\texttt{cow laying down because of being tired}'' in Figure~\ref{fig:vqavisu}(b). A large portion of VQAv2 questions is about detailed image contents, such as the object color in Figures~\ref{fig:vqavisu}(c-e). In the success cases,~\oursmodel~answers such questions with relevant textual descriptions if available, or by guessing via object properties. For example,  the description ``\texttt{a sliver refrigerator}'' in Figure~\ref{fig:vqavisu}(c) and the guess ``\texttt{bedroom walls are usually white}'' in Figure~\ref{fig:vqavisu}(d). However, by only looking at the incomplete textual description of the image,~\oursmodel~does fail on many questions. For example, it fails to predict the correct color in Figure~\ref{fig:vqavisu}(e) and the number of giraffes in Figure~\ref{fig:vqavisu}(f). We expect an end-to-end vision encoder tuning can help answer such questions better.
\section{Conclusion}
We present \oursmodel, an approach that uses GPT-3 for few-shot knowledge-based VQA. Instead of using explicit structured knowledge bases to retrieve and reason external knowledge, \oursmodel~jointly acquires and processes relevant knowledge by prompting GPT-3. It inherits GPT-3's strong few-shot ability, and surpasses the supervised state of the art on OK-VQA by a significant margin. Analyses show that our method implicitly acquires relevant knowledge to answer the question.

{
\small
\bibliography{egbib} 
}
\clearpage
\appendix
\twocolumn[{
\begin{center}
\Large 
\textbf{An Empirical Study of GPT-3 for Few-Shot Knowledge-Based VQA}\\(Supplementary Material)
\vspace{2em}

    \centering
    \includegraphics[width=17.5cm]{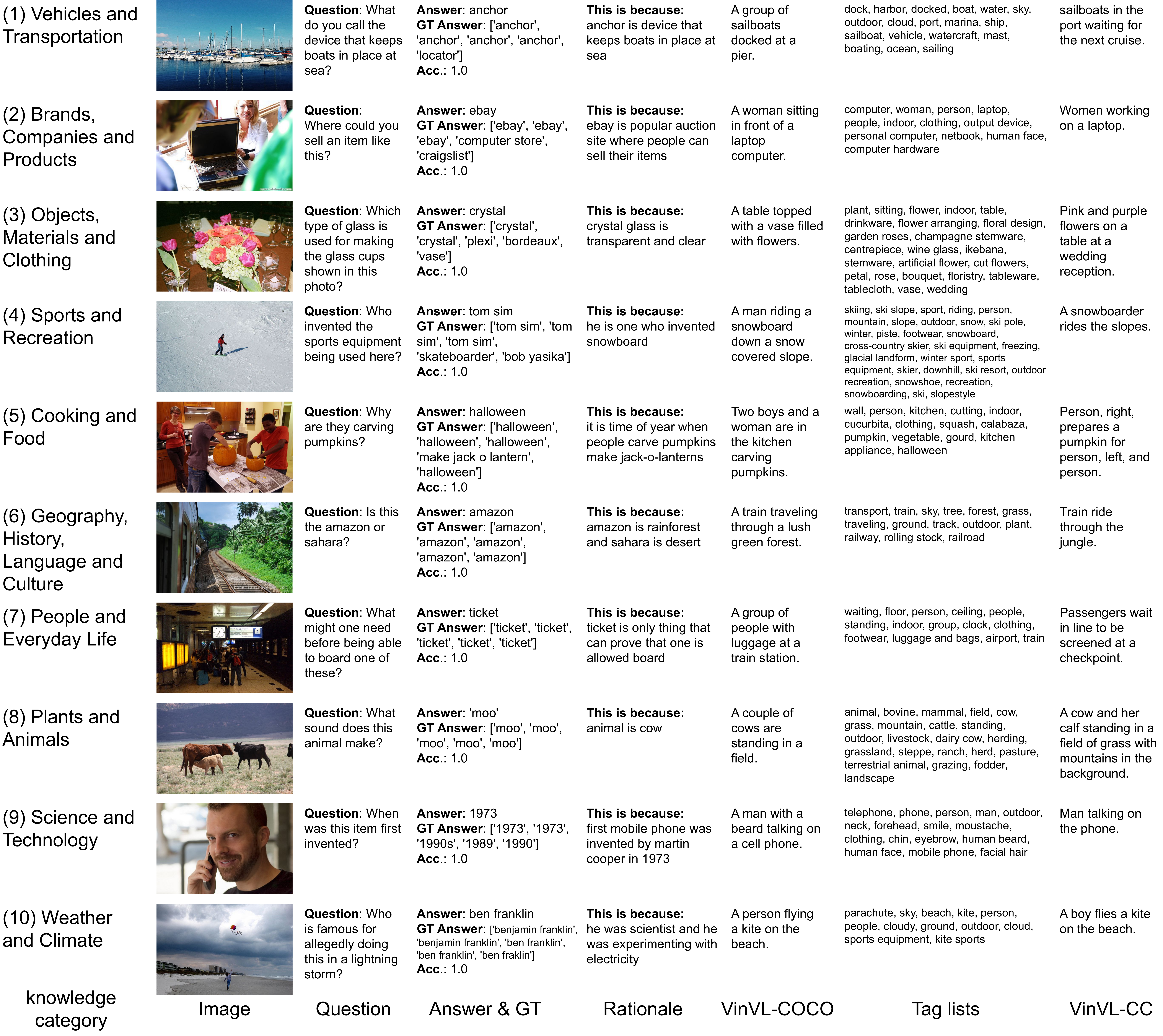}
    \vspace{-0.1in}
    \captionof{figure}{\small Additional qualitative results of our proposed \textbf{\texttt{PICa}} method on the OK-VQA dataset.}
    \label{fig:visusupp}
\vspace{1em}
\end{center}%
}]

We present additional qualitative results of our \textbf{\texttt{PICa}} method in Figure~\ref{fig:visusupp}. Same as in the main paper's Figure~\ref{fig:okvisu}, we present the results of \textbf{\texttt{PICa-Full}} with VinVL-COCO captions and tags, which has an accuracy of $48.0\%$. In the left four columns, we show the question knowledge category, the VQA input image, question, and predicted/ground-truth answers. We also include the answer rationale generated by GPT-3. In the right three columns, we include examples of the translated textual descriptions, \ie, ``VinVL-Caption-COCO'', ``Tag lists'', and ``VinVL-Caption-CC''.
We include one example from each of the $10$ knowledge categories defined by OK-VQA~\cite{marino2019ok}, as shown in rows (1-10). We observe that \textbf{\texttt{PICa}} performs uniformly well on all knowledge categories.

We highlight the following examples.
\vspace{-2pt}
\begin{itemize} 
\setlength\itemsep{-2pt}
\item \textbf{\texttt{PICa}} is strong in implicitly retrieving the encyclopedia knowledge to answer relevant questions. For example, the knowledge ``\texttt{tom sim invented snowboard}''\footnote{\url{www.google.com/search?q=who+invented+snowboard}} in row (4) and ``\texttt{martin cooper invented mobile phone in 1973}''\footnote{\url{www.google.com/search?q=when+was+mobile+phone+invented}} in row (9). Meanwhile, it also works well on questions that require other knowledge types, such as ``\texttt{cow makes sound moo}'' in row (8). The diverse knowledge in GPT-3 facilitates the use of a single implicit knowledge resource to replace the previous multiple explicit knowledge bases.
\item \textbf{\texttt{PICa}} jointly retrieves and reasons the knowledge implicitly by prompting GPT-3. Such joint modeling also helps to answer questions that require further reasoning beyond using a single piece of knowledge. For example, in row (5), ``\texttt{people are carving pumpkins to make jack o lantern}'', which is a tradition for ``\texttt{Halloween}''. Another example is in row (10), where \textbf{\texttt{PICa}} knows that ``\texttt{Ben Franklin}'' is famous for ``\texttt{flying kite in lightning storm}'', and ``\texttt{he is experimenting with electricity}''.
\end{itemize}

\end{document}